\documentclass[twocolumn]{rps-esrel2023}
\usepackage{xcolor}

\def\papername{\jobname}

\usepackage[authoryear]{natbib}
\usepackage{url}
\usepackage{subcaption}
\usepackage{graphicx}
\usepackage{float}

\graphicspath{}

\begin{document}


\twocolumn[

\title{Predictive Maintenance Model Based on Anomaly Detection in Induction Motors: A Machine Learning Approach Using Real-Time IoT Data}
\author{Sergio F. Chevtchenko $^{1}$, Monalisa. C. M. dos Santos $^{1}$, Diego. M. Vieira $^{1}$, Ricardo. L. Mota $^{1}$, Elisson Rocha $^{1}$, Bruna. V. Cruz $^{1}$, Danilo Araújo$^{2}$, Ermeson Andrade$^{2}$}
\address{\{sergio.chevtchenko, monalisa.santos, diego.vieira, ricardo.mota, elisson.rocha, bruna.cruz, \}@sistemafiepe.org.br, \{danilo.araujo, ermeson.andrade\}@ufrpe.br}

\address{1 ISI-TICs – SENAI Innovation Institute for Information and Communication Technologies}
\address{2 UFRPE, Departamento de Computação}

\begin{abstract} {With the support of Internet of Things (IoT) devices, it is possible to acquire data from degradation phenomena and design data-driven models to perform anomaly detection in industrial equipment. 
This approach not only identifies potential anomalies but can also serve as a first step toward building predictive maintenance policies. 
In this work, we demonstrate a novel anomaly detection system on induction motors used in pumps, compressors, fans, and other industrial machines. 
This work evaluates a combination of pre-processing techniques and machine learning (ML) models with a low computational cost. 
We use a combination of pre-processing techniques such as Fast Fourier Transform (FFT), Wavelet Transform (WT), and binning, which are well-known approaches for extracting features from raw data. 
We also aim to guarantee an optimal balance between multiple conflicting parameters, such as anomaly detection rate, false positive rate, and inference speed of the solution. To this end, multiobjective optimization and analysis are performed on the evaluated models. Pareto-optimal solutions are presented to select which models have the best results regarding classification metrics and computational effort. Differently from most works in this field that use publicly available datasets to validate their models, we propose an end-to-end solution combining low-cost and readily available IoT sensors. The approach is validated by acquiring a custom dataset from induction motors. Also, we fuse vibration, temperature, and noise data from these sensors as the input to the proposed ML model. Therefore, we aim to propose a methodology general enough to be applied in different industrial contexts in the future.}
\end{abstract}

\keywords{Anomaly detection, induction motors, multiobjective, machine learning, IoT, computational effort. }]

\section{Introduction}\label{sec1}
Industrial machines are fundamental assets in many contexts, and their proper functioning is crucial to maintain production efficiency and avoiding costly downtimes. Prognostics and Health Management (PHM) is a multidisciplinary field that focuses on developing methodologies and tools to monitor and diagnose the health of machines by predicting anomalous behavior, preventing failures, and improving reliability, safety, and performance~\citep{nguyen2022review}. 
With the advent of the IoT, it is possible to acquire data from sensors and devices in industrial environments (e.g. vibration, temperature, and acoustic sensors). This data can be used to design data-driven models that perform anomaly detection of industrial equipment~\citep{gan2020prognostics}.

Artificial Intelligence (AI) has emerged as a powerful tool for solving complex problems. The field of AI encompasses a wide range of techniques and algorithms that enable machines to learn, reason, and make decisions based on data~\citep{wang2019artificial}. AI techniques such as machine learning (ML), and deep learning (DL) are used currently in many fields, such as healthcare, finance, and manufacturing~\citep{chien2020artificial, goodell2021artificial}. In particular, AI has proven to be a valuable asset in Predictive Maintenance (PdM) for industrial machines~\cite{luo2022multi}. By using monitoring data, AI can detect anomalies and potential issues within the machine, leading to proactive maintenance, reducing downtime, and improving availability.

In this work, we focus on anomaly detection in induction motors, which is a challenging task due to the complex and nonlinear nature of their degradation phenomena. This can lead to multiple fault modes and subtle changes in their operational behavior. 
Therefore, we propose a novel methodology based on ML pipelines considering a low computational cost, and online and real-time applications.  

Our approach uses FFT \citep{walker2017fast}, WT \citep{chen2016wavelet}, and binning as pre-processing techniques to extract relevant features from raw data~\citep{kuhn2013data}. 
Then, we use OCSVM~\citep{pang2022hybrid}, IF~\citep{elnour2020dual}, and Local Outlier Factor (LOF) as ML models to classify whether the system has an anomaly or not. To ensure an optimal balance between multiple conflicting parameters, such as anomaly detection rate, false positive rate, and computational cost of the solution, we perform multiobjective optimization~\citep{tian2021evolutionary} and analysis on the evaluated models. 
We present Pareto-optimal solutions to select which models have the best results regarding classification metrics and computational effort. 

Most of the related literature on anomaly detection is based on publicly available datasets that tend to have data acquired with costly and reliable sensors. In contrast, our work presents an end-to-end solution involving a combination of low-cost and readily available IoT sensors to acquire data from induction motors. 
The fusion of vibration, temperature, and noise data from these sensors is provided as input to the evaluated ML models. This approach is general enough for our methodology to be applied in different industrial contexts in the future.


This paper is organized as follows. Section \ref{sec_relates} provides an overview of related works. Section \ref{sec_proposed} shows the methodology used to deploy this study. Section \ref{sec_results} presents the results. Finally, Section \ref{sec_concls} concludes the study and provides future directions related to this investigation.



\section{Related Works}\label{sec_relates}


Concerning nonintegrated devices (i.e. separated sensors), \cite{yang2016feature} proposed a methodology based on multilayer feedforward networks (MFNs) to address broken-rotor-bar and bearing faults from induction motors using data from current, accelerometers, gyroscope, and microphone signals. \cite{delgado2017methodology} use Complete Emsemble Mode Decomposition (CEEMED) to decompose and analyze acoustic and vibration signals to detect faults in induction motors - in this work, they use an acoustic microphone and an accelerometer as sensors. Our work is based on a non-invasive device to monitor the machine's condition with vibration, temperature, and acoustic sensors condensed in one device.  

Using only acoustic signals, \cite{glowacz2018acoustic} propose a methodology for fault diagnosis following the traditional pipeline in this field - preprocessing the data, performing feature extraction, and classification - using a low-cost microphone and a digital voice recorder. In our work, however, we used an integrated sensor to capture acoustic signals.


\cite{li2019deep} use kernel-based support vector machine to detect anomalies in the data collected from mechanical equipment. The vibration signals are measured through accelerometers at different rotating speed. The present work evaluates a combination of sensors, and the signal is analyzed and processed using Wavelet Transform (WT), and FFT. \cite{glowacz2019detection} use Method of Selection of Amplitudes of Frequencies (MSAF-12), as well as FFT and mean vector sum to perform feature extraction of vibration signals and detect deterioration on the rotor bar in motors using low-cost accelerometers.


To detect fault occurrence in real time, an online anomaly detection method with streaming data is proposed by \cite{liu2021online}, based on fine-grained feature forecasting. An unsupervised online anomaly detection model is used to analyze the vibration signals, as well as a fault alarm strategy is purposed for the prediction of fine-grained features. The present work is also aimed at real time anomaly detection with fusion of accelerometer, gyroscope, and microphone signals. 


In order to detect faults in complex electromechanical equipment, a multi-mode non-Gaussian variational autoencoder (MNVAE) is used to analyze the vibration signals in \cite{luo2022multi}. The OCSVM and other deep learning approaches are compared to prove the superiority of MNVAE to other methods. Due to the lower computational cost, the current work is focused on the evaluation of traditional ML algorithms for anomaly detection. 
The evaluation of deep learning models, including but not limited to Convolutional Neural Networks (CNNs) and Variational Autoencoders(VAEs), remains a topic for future investigation.

\section{Proposed Methodology}\label{sec_proposed}

\begin{figure*}[!htbp]
    \centering
    \includegraphics[scale=0.26]{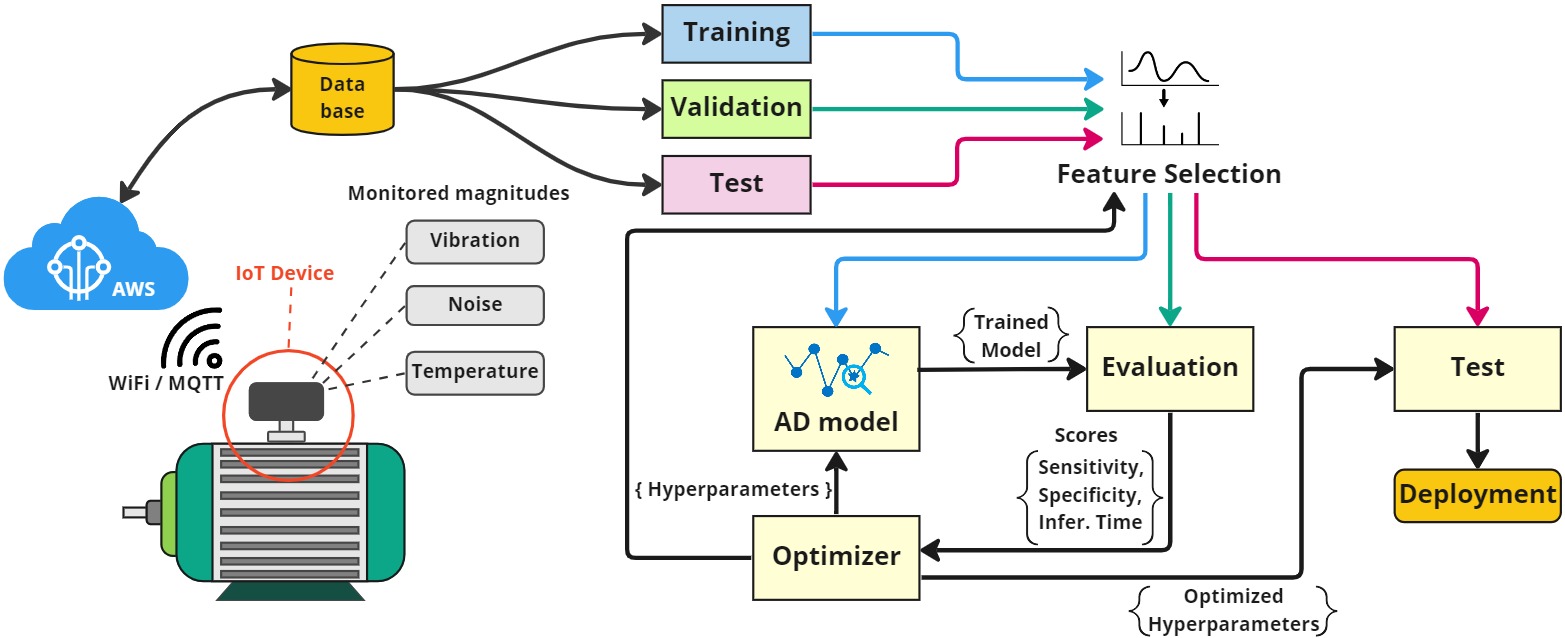}
    \caption{An overview of the proposed methodology.} 
    \label{fig_RMM_schema}
\end{figure*}

The proposed method combines an IoT sensor using off-the-shelf components with a multiobjective optimization of several well-known pre-processing and anomaly detection algorithms. Overall, the aim of this method is to produce an optimal configuration of hyperparameters in terms of both detection accuracy and processing speed. An overview of this methodology is presented in Figure \ref{fig_RMM_schema}.

\subsection{IoT Sensors}\label{ssec_proposed_iot}

As part of the current research, we develop an easy-to-install IoT device composed of low-cost sensors that collect data to be analyzed by the anomaly detection algorithm. This device has a neodymium magnet in its structure that is strong enough to fix it on any steel surface. This device collects the temperature, vibration, and noise magnitudes of the machine through an integrated set of sensors with a defined frequency. The collected data is structured and sent via MQTT to a cloud messaging service, as illustrated in Figure \ref{fig_RMM_schema}. This service is responsible for storing the messages with the collected data until they are consumed by the anomaly detection model.




The control unit of the device is composed of an ESP32 processor module and the sensor elements are composed of  a 6-axis vibration sensor, 2 noise sensors, and a temperature sensor.



The vibration sensor ISM330DLCTR has a 3-axis accelerometer and a 3-axis gyroscope with a sampling rate of 1kHz. The noise sensor consists of 2 identical microphones SPH0645LM4H-B, positioned internally on opposite sides of the device's casing. One microphone is at the bottom and the other one at the top, both operate at a sampling rate of 20kHz. The temperature is measured with the analog sensor MCP9700T-E/TT in the range between -40ºC and 150ºC and is then converted to a digital signal. The final temperature reading is an average of 100 samples. 

\subsection{Preprocessing}\label{ssec_proposed_preproc}

Each sample from the temperature sensor, accelerometer, gyroscope, and microphone is acquired by the IoT device and stored in a database. With the exception of temperature, the samples are multidimensional vectors. Thus, the next step of data preparation aims at retrieving the most relevant information from any given data packet. In this work, we consider three techniques for preprocessing our data: Fast Fourier Transform (FFT); Discrete Wavelet Transform (DWT); and Binning.

\subsection{Dimensionality Reduction}\label{ssec_proposed_dim_red}

Principal Component Analysis (PCA) is a popular technique used for dimensionality reduction in various fields, but mostly in data science and machine learning. PCA aims to capture the maximum amount of variation in the data with a smaller number of variables or dimensions, which preserves most of the variance by retaining the components with the highest eigenvalues. PCA can identify and eliminate redundant or highly correlated variables, this allows it to reduce data redundancy. Also, models built on PCA-transformed data are often faster and more efficient than those built in original data. By transforming the data into a lower-dimensional space, PCA makes it easier to visualize and interpret the relationships between variables. In addition, PCA generalizes well, because it can be applied to a wide range of data types. Overall, PCA is a powerful technique for reducing the dimensionality of data while retaining most of the relevant information.  

\subsection{Anomaly Detection}\label{ssec_proposed_anom}

Once enough normal operation data is collected, an anomaly detection algorithm can be trained. Note that the training is performed exclusively on the normal data and thus the algorithm is expected to identify future, and possibly rare, anomalous observations based on its knowledge of normal data patterns. Three commonly used anomaly detection algorithms are considered in the present work: One-class SVM (Support Vector Machine); Isolation Forest (IF); and Local Outlier Factor (LOF). We implemented these models using the sci-kit learn library~\citep{scikit_learn} - the chosen ML models are described below.

One-class SVM is a machine learning technique used to identify unusual or anomalous data points in a given dataset. One-class SVM requires only normal training data to build a model that can distinguish between normal and anomalous data points. In the current implementation, the following hyperparameters are considered: Kernel - specifies the kernel type to be used in the algorithm (linear, 3rd order polynomial or RBF); Nu - an upper bound on the fraction of training errors and a lower bound of the fraction of support vectors,  from [0.25, 0.5, 0.75]; Tolerance - a stopping criterion, from [$10^{-4}$, $10^{-3}$, $10^{-2}$]. 

The Isolation Forest works by constructing a random forest of decision trees. The isolation score is calculated based on the number of splits needed to isolate the anomaly, and it is normalized by the average path length of unsuccessful splits in the tree. In the present work, the IF is implemented using two parameters: the n\_estimators (the number of base estimators in the ensemble) and the max\_samples (the number of samples to draw from X to train each base estimator). The main advantage of Isolation Forest is it requires less memory and computational resources. 

LOF measures the local density around a data point by comparing its distance to its k-nearest neighbors with the average distance of those neighbors to each other. If the distance of a data point to its k-nearest neighbors is significantly smaller than the average distances of those neighbors to each other, then the data point is considered to be in a dense region and has a low LOF value. On the other hand, if the distance is significantly larger, then the data point is considered to be in a sparse region and has a high LOF value. The LOF algorithm assigns anomaly scores to each point based on its LOF value, with higher scores indicating more anomalous data points. The n\_neighbors (number of neighbors to use by default for neighbors queries) parameter is used for tuning the LOF algorithm by the optimization step.

\subsection{Optimization}\label{ssec_proposed_optim}

The performance of the previously presented pre-processing and anomaly detection algorithms can be significantly influenced by the selection of appropriate hyperparameters. Thus, in order to avoid manual tuning, an automatic optimization approach is adopted. Given a model with a corresponding set of hyperparameters, the following three metrics are considered: \textbf{Sensitivity} -- also known as the true positive rate or recall, sensitivity is the proportion of actual anomalous cases that are correctly identified by the model as anomalous; \textbf{Specificity} -- the proportion of actual normal samples that are correctly identified by the model as normal; \textbf{Inference time} -- this is measured as the time elapsed from pre-processing to when the model produces a binary classification, averaged across all samples in the validation subset. 

Due to the multiobjective nature of the above optimization problem, NSGA-II algorithm \citep{deb2002fast} is used to iteratively search for Pareto-optimal configurations. We use the default configuration of NSGA-II, provided by the Optuna framework \citep{akiba2019optuna}.

\section{Experimental Results}\label{sec_results}

\subsection{Setup}\label{ssec_setup}

\subsubsection{The Motor Anomaly Dataset}\label{sssec_setup_dataset}
\begin{figure}[!htbp]
    \centering
    \includegraphics[width=0.8 \linewidth]{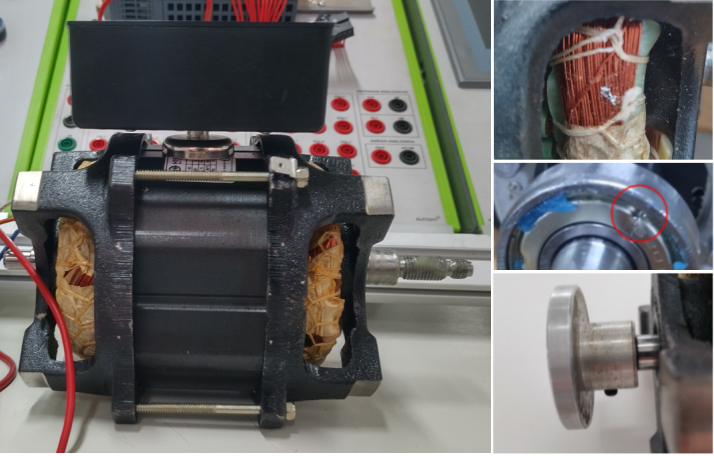}
    \caption{The IoT sensor attached to an electric motor setup (left). Short-circuited stator, damaged bearing and an unbalanced shaft load (right).}
    \label{fig_motor_inverter}
\end{figure}

The dataset is collected using the IoT sensor, described in Section \ref{ssec_proposed_iot}. The sensor is magnetically attached to a 1 hp three-phase induction motor, driven by an inverter at a constant speed of 1000 rpm, as illustrated in Figure \ref{fig_motor_inverter}. 

Induction motors are among the most common type of devices in the industrial environment. In the present work, three distinct faults are reproduced to generate anomalous data. The following failures are also among the most commonly encountered in the field~\citep{gundewar2021condition}:

\begin{itemize}
    \item Unbalanced load (Motor 1) -- a small unbalanced load is placed on the motor shaft; 
    \item Damaged bearings (Motor 2) -- the metallic seal of the bearing is punctured without causing the bearings to stall;
   \item Stator short circuit (Motor 3) -- an inter-turn short circuit is caused in the stator winding.
\end{itemize}

Note that the anomalies above are reproduced with the aim of being hard to detect during the operation of a machine. This is because the change in overall noise and vibration levels is low when compared to normal data. 

The same amount of data is acquired for each of the three motors and partitioned into the following subsets:
\begin{itemize}
    \item \textit{Training} -- 500 consecutive samples of the normal operation;
    \item \textit{Validation} -- 500 samples of normal operation and 500 samples of anomalous data; 
    \item \textit{Test} -- 250 samples of normal operation and 250 samples of anomalous data.
\end{itemize}
Thus, considering the three motors, the dataset is composed of 3750 normal and 2250 anomaly samples, collected in approximately 30s intervals. The dataset is available upon request.

\subsubsection{Multi-objective Optimization}\label{sssec_setup_mo_opt}
Given an anomaly detection algorithm (OC-SVM, IF, or LOF) and the \textit{Training} and \textit{Validation} subsets, the optimization loop is tasked with finding the best set of hyperparameters that provide Pareto-optimal solutions to the three conflicting objectives described in Section \ref{ssec_proposed_optim}. In order to provide consistent results in terms of inference time, all algorithms are evaluated on a single core of an i5 CPU.

\begin{figure}[!htbp]
    \centering
    \includegraphics[width=0.9 \linewidth]{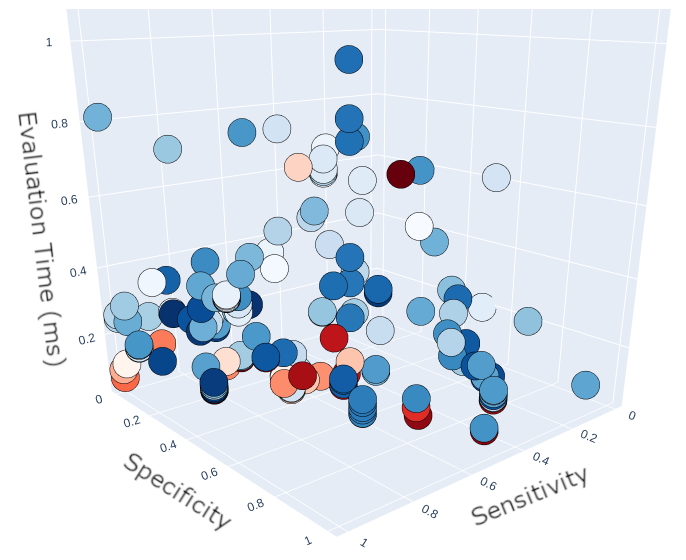}
    \caption{Trials within an optimization run.}
    \label{fig_optuna_sample}
\end{figure}

It is worth noting that, differently from the anomaly detection models, the optimization algorithm can be viewed as supervised training. This is because the \textit{Validation} subset contains an equal amount of normal and anomaly data. The hyperparameter search space is composed of the individual parameters described in Sections \ref{ssec_proposed_preproc}, \ref{ssec_proposed_dim_red}, and \ref{ssec_proposed_anom}. The NSGA-II optimizer is given an evaluation budget of 200 trials and the configuration closest to the optimal one is tested in the \textit{Test} subset. An optimization run consisting of 100 trials is illustrated in Figure \ref{fig_optuna_sample}, the warmer colors correspond to non-dominated solutions. 

\subsection{Results and Discussion}\label{ssec_results}
For each of the three anomaly detection algorithms, ten of the best configurations are selected for evaluation on the \textit{Test} subset. The selection is based on proximity to the optimal sensitivity and specificity.  Because of the possibility of overfitting during optimization, this is done in order to evaluate the overall validity of the proposed approach, including the multiobjective optimization. 

\begin{table*}
\centering
\caption{A summary of the average evaluation results on the \textit{validation}~/~\textit{test} subsets}
\label{table_res_val_test}
\begin{tabular}{c c c c}
\hline
\textbf{Algorithm} & \begin{tabular}{c} \textbf{Sensitivity} \\ \textbf{(\%)} \end{tabular} & \begin{tabular}{c} \textbf{Specificity} \\ \textbf{(\%)} \end{tabular} & \begin{tabular}{c} \textbf{Inference time} \\ \textbf{(ms)} \end{tabular}\\
\hline
OC-SVM  & 73.1~/~47.9 & 63.4~/~36.5 & 0.43\\
IF & 88.9~/~86.3 & 61.6~/~67.4 &  21.40\\
LOF & 90.8~/~77.6 & 74.5~/~72.1 &  0.81\\
\hline
\end{tabular}
\end{table*}

\begin{figure}[!htbp]
    \centering
    \includegraphics[width=1.0 \linewidth]{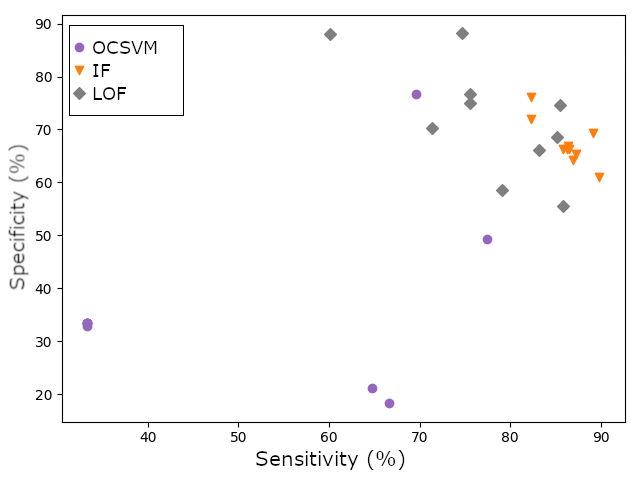}
    \caption{A comparison of sensitivity and specificity on the \textit{test} subset. 
    }
    \label{fig_sens_vs_spec_test}
\end{figure}

A comparison of the three optimized algorithms in terms of sensitivity and specificity is presented in Figure \ref{fig_sens_vs_spec_test}. Each dot of the same color represents one of the best trials, selected during optimization. Isolation Forest and Local Outlier Factors provide the best results on this trade-off.

An additional comparison, including \textit{validation} subset and inference time, is presented in Table \ref{table_res_val_test}. The results indicate that the LOF algorithm obtains the best average performance in terms of the three objectives. The best-ranked configuration of this algorithm uses FFT with 10 bins for processing of noise and vibration signals, followed by binning using 10 bins and PCA with two principal components. The anomaly detection is performed using 200 neighbors. While the above parameters are unique to this configuration, it is worth noting that the ten best sets of hyperparameters have some common attributes. First, FFT is always present, suggesting that WDT transform does not offer an advantage when combined with LOF. Second, PCA reduction is always used and in 8 out of 10 configurations only two principal components are necessary. 

\begin{figure*}[!htbp]
    \centering
    \includegraphics[width=0.7 \linewidth]{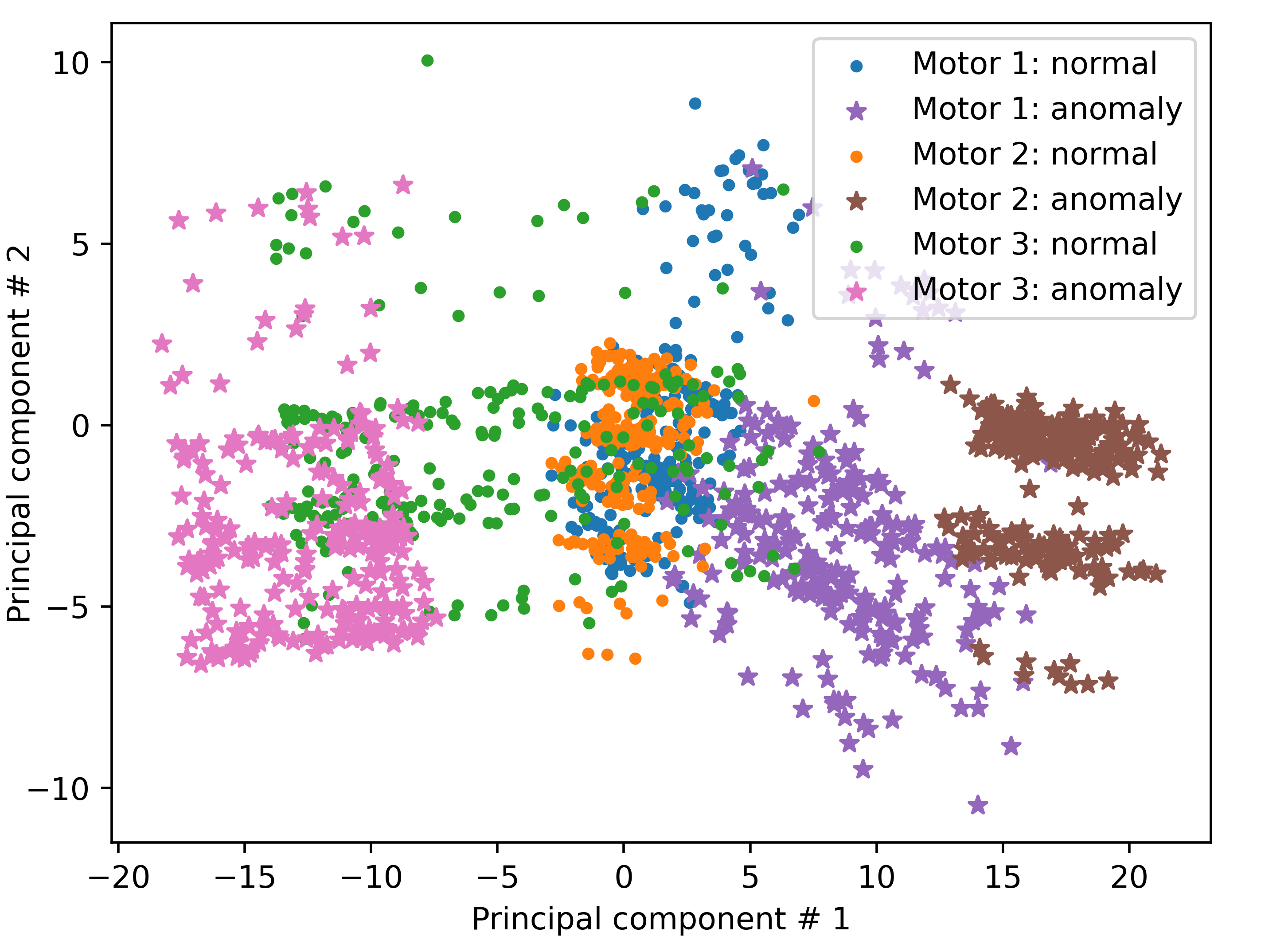}
    \caption{The \textit{test} subset using PCA transform.}
    \label{fig_pca_plot_test}
\end{figure*}

The \textit{test} subset can be visualized using the optimized pre-processing strategy, as illustrated in Figure \ref{fig_pca_plot_test}. Interestingly, Motor 3 has a different normal signature from motors 1 and 2. This is probably because the first two motors come from the same new batch, while the third motor was previously in operation for approximately a year. The different clusters formed by the anomaly data are expected and reflect the different nature of the anomaly, as described in Section \ref{sssec_setup_dataset}.

\section{Conclusion and Future Works}\label{sec_concls}
An end-to-end anomaly detection solution is presented and evaluated. The proposed approach includes parametrized data preprocessing and dimensionality reduction, as well as the selection of hyperparameters of three evaluated anomaly detection algorithms: OC-SVM, IF, and LOF. 

A new induction motor anomaly dataset is also introduced. The dataset contains 3750 normal and 2250 anomaly samples, divided into three commonly found defects: unbalanced load, damaged bearings, and stator short circuits. The defects are implemented in a way that does not significantly affect the performance of the motors and would be difficult to spot on a factory floor. 

The experimental results suggest that LOF is a promising algorithm for future evaluation in a real-world scenario, considering sensitivity, specificity, and inference time metrics.   

Future endeavors could explore several directions, such as the evaluation of an ensemble of models to reduce overfitting observed during optimization. Deep learning models, such as VAEs should also be evaluated using the proposed multiobjective approach. Finally, we intend to increase the diversity and size of the dataset with additional real-world machines.

\begin{acknowledgement}
The authors would like to thank ISI-TICs and the 'Brazilian Agency for Industrial Development - ABDI' for financial support. We also would like to thank 'Foundation for Technological Innovations - FITec' for their technological support.

\end{acknowledgement}
\bibliographystyle{chicago}

\bibliography{Reference}

\begin{thebibliography}{}

\bibitem[\protect\citeauthoryear{Akiba, Sano, Yanase, Ohta, and Koyama}{Akiba
  et~al.}{2019}]{akiba2019optuna}
Akiba, T., S.~Sano, T.~Yanase, T.~Ohta, and M.~Koyama (2019).
\newblock Optuna: A next-generation hyperparameter optimization framework.
\newblock In {\em Proceedings of the 25th ACM SIGKDD international conference
  on knowledge discovery \& data mining}, pp.\  2623--2631.

\bibitem[\protect\citeauthoryear{Chen, Li, Pan, Chen, Zi, Yuan, Chen, and
  He}{Chen et~al.}{2016}]{chen2016wavelet}
Chen, J., Z.~Li, J.~Pan, G.~Chen, Y.~Zi, J.~Yuan, B.~Chen, and Z.~He (2016).
\newblock Wavelet transform based on inner product in fault diagnosis of
  rotating machinery: A review.
\newblock {\em Mechanical systems and signal processing\/}~{\em 70}, 1--35.

\bibitem[\protect\citeauthoryear{Chien, Dauz{\`e}re-P{\'e}r{\`e}s, Huh, Jang,
  and Morrison}{Chien et~al.}{2020}]{chien2020artificial}
Chien, C.-F., S.~Dauz{\`e}re-P{\'e}r{\`e}s, W.~T. Huh, Y.~J. Jang, and J.~R.
  Morrison (2020).
\newblock Artificial intelligence in manufacturing and logistics systems:
  algorithms, applications, and case studies.

\bibitem[\protect\citeauthoryear{Deb, Pratap, Agarwal, and Meyarivan}{Deb
  et~al.}{2002}]{deb2002fast}
Deb, K., A.~Pratap, S.~Agarwal, and T.~Meyarivan (2002).
\newblock A fast and elitist multiobjective genetic algorithm: Nsga-ii.
\newblock {\em IEEE transactions on evolutionary computation\/}~{\em 6\/}(2),
  182--197.

\bibitem[\protect\citeauthoryear{Delgado-Arredondo, Morinigo-Sotelo,
  Osornio-Rios, Avina-Cervantes, Rostro-Gonzalez, and de~Jesus
  Romero-Troncoso}{Delgado-Arredondo et~al.}{2017}]{delgado2017methodology}
Delgado-Arredondo, P.~A., D.~Morinigo-Sotelo, R.~A. Osornio-Rios, J.~G.
  Avina-Cervantes, H.~Rostro-Gonzalez, and R.~de~Jesus Romero-Troncoso (2017).
\newblock Methodology for fault detection in induction motors via sound and
  vibration signals.
\newblock {\em Mechanical Systems and Signal Processing\/}~{\em 83}, 568--589.

\bibitem[\protect\citeauthoryear{Elnour, Meskin, Khan, and Jain}{Elnour
  et~al.}{2020}]{elnour2020dual}
Elnour, M., N.~Meskin, K.~Khan, and R.~Jain (2020).
\newblock A dual-isolation-forests-based attack detection framework for
  industrial control systems.
\newblock {\em IEEE Access\/}~{\em 8}, 36639--36651.

\bibitem[\protect\citeauthoryear{Gan}{Gan}{2020}]{gan2020prognostics}
Gan, C.~L. (2020).
\newblock Prognostics and health management of electronics: Fundamentals,
  machine learning, and the internet of things: John wiley \& sons ltd.(2018).
  pp. 731.
\newblock {\em Life Cycle Reliability and Safety Engineering\/}~{\em 9\/}(2),
  225--226.

\bibitem[\protect\citeauthoryear{Glowacz}{Glowacz}{2018}]{glowacz2018acoustic}
Glowacz, A. (2018).
\newblock Acoustic based fault diagnosis of three-phase induction motor.
\newblock {\em Applied Acoustics\/}~{\em 137}, 82--89.

\bibitem[\protect\citeauthoryear{Glowacz, Glowacz, Kozik, Piech, Gutten,
  Caesarendra, Liu, Brumercik, Irfan, and Khan}{Glowacz
  et~al.}{2019}]{glowacz2019detection}
Glowacz, A., W.~Glowacz, J.~Kozik, K.~Piech, M.~Gutten, W.~Caesarendra, H.~Liu,
  F.~Brumercik, M.~Irfan, and Z.~F. Khan (2019).
\newblock Detection of deterioration of three-phase induction motor using
  vibration signals.
\newblock {\em Measurement Science Review\/}~{\em 19\/}(6), 241--249.

\bibitem[\protect\citeauthoryear{Goodell, Kumar, Lim, and Pattnaik}{Goodell
  et~al.}{2021}]{goodell2021artificial}
Goodell, J.~W., S.~Kumar, W.~M. Lim, and D.~Pattnaik (2021).
\newblock Artificial intelligence and machine learning in finance: Identifying
  foundations, themes, and research clusters from bibliometric analysis.
\newblock {\em Journal of Behavioral and Experimental Finance\/}~{\em 32},
  100577.

\bibitem[\protect\citeauthoryear{Gundewar and Kane}{Gundewar and
  Kane}{2021}]{gundewar2021condition}
Gundewar, S.~K. and P.~V. Kane (2021).
\newblock Condition monitoring and fault diagnosis of induction motor.
\newblock {\em Journal of Vibration Engineering \& Technologies\/}~{\em 9},
  643--674.

\bibitem[\protect\citeauthoryear{Kuhn, Johnson, Kuhn, and Johnson}{Kuhn
  et~al.}{2013}]{kuhn2013data}
Kuhn, M., K.~Johnson, M.~Kuhn, and K.~Johnson (2013).
\newblock Data pre-processing.
\newblock {\em Applied predictive modeling\/}, 27--59.

\bibitem[\protect\citeauthoryear{Li, Li, Wang, and Wang}{Li
  et~al.}{2019}]{li2019deep}
Li, Z., J.~Li, Y.~Wang, and K.~Wang (2019).
\newblock A deep learning approach for anomaly detection based on sae and lstm
  in mechanical equipment.
\newblock {\em The International Journal of Advanced Manufacturing
  Technology\/}~{\em 103}, 499--510.

\bibitem[\protect\citeauthoryear{Liu, Mao, Shi, Wu, and Chen}{Liu
  et~al.}{2021}]{liu2021online}
Liu, K., W.~Mao, H.~Shi, C.~Wu, and J.~Chen (2021).
\newblock Online anomaly detection with streaming data based on fine-grained
  feature forecasting.
\newblock In {\em 2021 33rd Chinese Control and Decision Conference (CCDC)},
  pp.\  454--459. IEEE.

\bibitem[\protect\citeauthoryear{Luo, Chen, Zi, Chang, and Feng}{Luo
  et~al.}{2022}]{luo2022multi}
Luo, Q., J.~Chen, Y.~Zi, Y.~Chang, and Y.~Feng (2022).
\newblock Multi-mode non-gaussian variational autoencoder network with missing
  sources for anomaly detection of complex electromechanical equipment.
\newblock {\em ISA transactions\/}.

\bibitem[\protect\citeauthoryear{Nguyen, Medjaher, and Tran}{Nguyen
  et~al.}{2022}]{nguyen2022review}
Nguyen, K.~T., K.~Medjaher, and D.~T. Tran (2022).
\newblock A review of artificial intelligence methods for engineering
  prognostics and health management with implementation guidelines.
\newblock {\em Artificial Intelligence Review\/}, 1--51.

\bibitem[\protect\citeauthoryear{Pang, Pu, and Li}{Pang
  et~al.}{2022}]{pang2022hybrid}
Pang, J., X.~Pu, and C.~Li (2022).
\newblock A hybrid algorithm incorporating vector quantization and one-class
  support vector machine for industrial anomaly detection.
\newblock {\em IEEE Transactions on Industrial Informatics\/}~{\em 18\/}(12),
  8786--8796.

\bibitem[\protect\citeauthoryear{Pedregosa, Varoquaux, Gramfort, Michel,
  Thirion, Grisel, Blondel, Prettenhofer, Weiss, Dubourg, Vanderplas, Passos,
  Cournapeau, Brucher, Perrot, and Duchesnay}{Pedregosa
  et~al.}{2011}]{scikit_learn}
Pedregosa, F., G.~Varoquaux, A.~Gramfort, V.~Michel, B.~Thirion, O.~Grisel,
  M.~Blondel, P.~Prettenhofer, R.~Weiss, V.~Dubourg, J.~Vanderplas, A.~Passos,
  D.~Cournapeau, M.~Brucher, M.~Perrot, and E.~Duchesnay (2011).
\newblock Scikit-learn: Machine learning in {P}ython.
\newblock {\em Journal of Machine Learning Research\/}~{\em 12}, 2825--2830.

\bibitem[\protect\citeauthoryear{Tian, Si, Zhang, Cheng, He, Tan, and Jin}{Tian
  et~al.}{2021}]{tian2021evolutionary}
Tian, Y., L.~Si, X.~Zhang, R.~Cheng, C.~He, K.~C. Tan, and Y.~Jin (2021).
\newblock Evolutionary large-scale multi-objective optimization: A survey.
\newblock {\em ACM Computing Surveys (CSUR)\/}~{\em 54\/}(8), 1--34.

\bibitem[\protect\citeauthoryear{Walker}{Walker}{2017}]{walker2017fast}
Walker, J.~S. (2017).
\newblock {\em Fast fourier transforms}.
\newblock CRC press.

\bibitem[\protect\citeauthoryear{Wang and Siau}{Wang and
  Siau}{2019}]{wang2019artificial}
Wang, W. and K.~Siau (2019).
\newblock Artificial intelligence, machine learning, automation, robotics,
  future of work and future of humanity: A review and research agenda.
\newblock {\em Journal of Database Management (JDM)\/}~{\em 30\/}(1), 61--79.

\bibitem[\protect\citeauthoryear{Yang, Pen, Wang, and Chang}{Yang
  et~al.}{2016}]{yang2016feature}
Yang, T., H.~Pen, Z.~Wang, and C.~S. Chang (2016).
\newblock Feature knowledge based fault detection of induction motors through
  the analysis of stator current data.
\newblock {\em IEEE Transactions on Instrumentation and Measurement\/}~{\em
  65\/}(3), 549--558.

\end{thebibliography}

\end{document}